\documentclass[10pt,twocolumn,letterpaper]{article}

\usepackage{cvpr}
\usepackage{times}
\usepackage{epsfig}
\usepackage{graphicx}
\usepackage{amsmath}
\usepackage{amssymb}

\usepackage{multirow}
\usepackage{bigstrut}
\usepackage{comment}
\usepackage{authblk}

\usepackage[breaklinks=true,bookmarks=false]{hyperref}

\cvprfinalcopy 

\def\cvprPaperID{****} 

\begin{document}

\title{Deep Attention Aware Feature Learning for Person Re-Identification}

\author{Yifan Chen\textsuperscript{1},\ Han Wang\textsuperscript{1},\ Xiaolu Sun\textsuperscript{2}, \ Bin Fan\textsuperscript{3}\thanks{\textit{Email:} \url{bfan@nlpr.ia.ac.cn}} ,\ Chu Tang\textsuperscript{2}\thanks{\textit{Email:} \url{chutang1022@gmail.com}} \\
	\tt\small{\textsuperscript{1}Tsinghua University, Beijing, China}  \\
	 \tt\small{\textsuperscript{2}Addx Technology, Beijing, China} \\
	 \tt\small{\textsuperscript{3}University of Science and Technology Beijing, Beijing, China}\\
}

\maketitle

\begin{abstract}
	Visual attention has proven to be effective in improving the performance of person re-identification. Most existing methods apply visual attention heuristically by learning an additional attention map to re-weight the feature maps for person re-identification. However, this kind of methods inevitably increase the model complexity and inference time. In this paper, we propose to incorporate the attention learning as additional objectives in a person ReID network without changing the original structure, thus maintain the same inference time and model size. Two kinds of attentions have been considered to make the learned feature maps being aware of the person and related body parts respectively. Globally, a holistic attention branch (HAB) makes the feature maps obtained by backbone focus on persons so as to alleviate the influence of background. Locally, a partial attention branch (PAB) makes the extracted features be decoupled into several groups and be separately responsible for different body parts (i.e., keypoints), thus increasing the robustness to pose variation and partial occlusion. These two kinds of attentions are universal and can be incorporated into existing ReID networks. We have tested its performance on two typical networks (TriNet~\cite{hermans2017defense} and Bag of Tricks~\cite{luo2019bag}) and observed significant performance improvement on five widely used datasets.\footnote{Our codes are publicly available at \url{https://github.com/CYFFF/DAAF_re-id}}
\end{abstract}


\section{Introduction}

\begin{figure}[t]		 
	\begin{center}
		\includegraphics[width=1\linewidth]{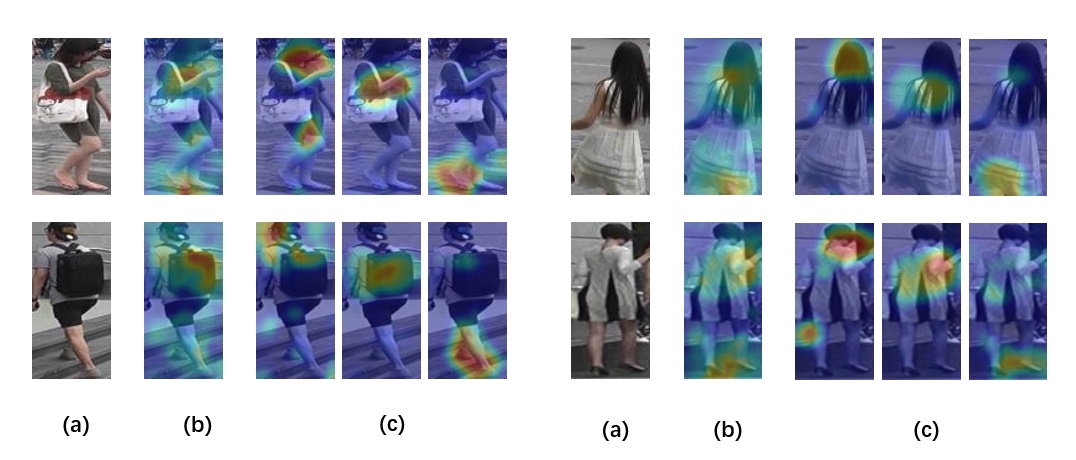}
	\end{center}
	\vspace{-4mm}
	\caption{An illustration of heatmaps generated by our attention aware feature learning. (a) original images; (b) holistic attention heatmaps; (c) partial attention heatmaps. Note that how our method effectively alleviates the background and makes the heatmaps focus on the person and the body parts, respectively.}
	\label{fig:fig1}
	\vspace{-5mm}
\end{figure}

Person re-identification (ReID) targets at matching the same person at different locations across different cameras. A common solution is to extract features directly from the whole person~\cite{hermans2017defense,sun2017svdnet}. For example, using a deep network pre-trained on ImageNet and fine-tuned over specific ReID datasets. Although the ReID problem has been extensively studied in recent years, it is still an open problem due to pose variation, background clutters, occlusion, etc.

To address these challenges, many solutions have been proposed in the literature. Among them, attention learning is attractive for its potential in removing background clutter \cite{song2018mask}, or enhancing local discriminability for different body parts \cite{zhao2017deeply}. Currently, most strategies using attention have to incorporate a separate stream as attention function to re-weight the feature maps which will increase computational complexity and model size in turn. Alternatively, we consider a more practical way to incorporate attention without changing the basic ReID networks. Our key assumption is that such attention can be implicitly embedded in the feature maps that are used for extracting person appearance representations. If feature maps contain such information, then they could be used to predict some attention-related information subsequently.

Specifically, this paper proposes an attention aware feature learning method for ReID task. We believe that CNN with proper constraints can obtain attention itself because of its powerful non-linearity. Therefore, if we can add appropriate constraints in the train stage, the attention aware features could be obtained without adding additional structures as previous works did. To this end, we present a holistic attention branch (HAB) to introduce the global attention information in the learned feature maps and a partial attention branch (PAB) to generate local attention aware feature maps, see Figure \ref{fig:fig1} for example. By predicting the mask of a person, HAB is designed to restrict the backbone network focus on person bodies instead of the background. PAB further forces different feature channels focus on different body parts, by explicitly using different feature groups  to predict different keypoints. While channel wise attention has recently been explored for segmentation~\cite{fu2019dual}, to the best of our knowledge, it has not been used in ReID. Meanwhile, the channel wise attention in PAB is quite different from the one used in~\cite{fu2019dual} which is basically to re-weight the feature channels. For comparison, PAB interprets channel wise attention as a way to make specific channels focus on different spatial parts and implicitly achieves the decoupling of feature channels.  

The main contributions of this work can be summarized as: (1) We propose a method to incorporate attention in feature learning without changing any structure of the original model, simply by adding supervisions during training. In other words, our approach adds no extra computing complexity during inference. (2) Both global and local attentions have been considered simultaneously in our method to improve the ReID performance. Especially, the partial attention implemented by performing channel wise decoupling supplies a new perspective to attention learning for ReID. (3) Our method has been extensively evaluated on widely used benchmarks with ablation study to analyze how different modules in our method work.

\section{Related Work}

\subsection{Attention learning in person ReID}  Attention learning has recently been introduced to person ReID task\cite{zhao2017deeply,li2018harmonious,song2018mask,tay2019aanet,si2018dual} for obtaining more discriminative features due to its successful in other computer vision tasks like object detection\cite{chen2017sca} and image segmentation\cite{chen2016attention,fu2019dual}. The common strategy to use attention in ReID is to incorporate a separate stream of regional attention into a deep convolutional ReID model. For example, Zhao et al.\cite{zhao2017deeply} exploit the Spatial Transformer Network~\cite{jaderberg2015spatial} as hard attention to search discriminative parts. Li et al. \cite{li2018harmonious} propose a multi-granularity attention selection mechanism to address the problem caused by poor located bounding-boxes and noisy information at pixel level. Song et al. \cite{song2018mask} propose a method generating a pair of body-aware and background-aware spatial attention maps for background clutter removal. 

The methods mentioned above could be regarded as spatial-wise attention learning. Channel-wise attention are also introduced in this area by \cite{wang2018mancs, chen2019abd, chen2019self}  to decide which channel is more important. Wang et al.~\cite{wang2018mancs} incorporate channel-wise attention by using a series of three-dimensional masks , which could be obtained through multi-task learning, to re-weight the feature map not only on spatial-wise but also on channel wise. Chen et al.~\cite{chen2019abd} use channel attention module which could calculate correlation coefficient among different channels to realize attention mechanism on channel-wise. Due to the additional streams used for attention, all these methods achieve better performance at the cost model size as well as the inference time. On the contrary, our method achieves attention learning only in the training stage and does not alter the inference network.

\subsection{Local features in person ReID} 

Extracting the appearance representations of a whole person image has been well studied for person ReID, and encouraging performance is achieved in the past years with the fast development of CNNs~\cite{zheng2016person,sun2017svdnet,hermans2017defense,zhong2018camera}. However, these global features usually do not perform well in cases of pose variation, occlusions and missing parts. Methods based on local features have been proposed to handle such issues. The most common way to extract local features is through image dicing, which directly divides person image into horizontal stripes\cite{yi2014deep,cheng2016person,sun2018beyond} or grids\cite{li2014deepreid, ahmed2015improved}. Then the extracted local features are assembled to form person appearance representations.

Another solution is performing part-alignment based on either keypoints or body parts which are estimated by separate models. For example, Varior et al. \cite{varior2016siamese} applies affine transformation to align the same keypoints. Kalayeh et al. \cite{kalayeh2018human} extract local features for different human body structure ROIs (Region-Of-Interest) to obtain local features and concatenate them with global features. Zhao et al. \cite{zhao2017deeply} obtain part-alignment representations according to the body parts detection. 

All of these part-alignment methods require an additional model~(either for keypoints regression or for semantic parsing), which leads to extra computation complexity. Furthermore, since all the exiting local feature based methods have to concatenate many feature embeddings, either from different image regions or body parts, it finally results in a very high dimensional embedding which are less efficiency to be dealt with in practice. For comparison, the proposed PAB achieves local feature learning in a significantly different way. In essential, it implicitly extracts the locally spatial features by decoupling feature maps at channel dimension which has never been explored before. What is more, our method does not need additional models to detect local features, thus keeping high efficiency in inference time.

\subsection{ReID with human mask and keypoints} With the development of human mask and keypoint detection, there are many ReID methods proposed to use mask or keypoints for getting more robust feature representations by excluding the background clutter or conducting part-alignment when extracting features. By simultaneously predicting human mask and extracting feature representations, MGCAM~\cite{song2018mask} and SPReID~\cite{kalayeh2018human} have reported improved results with the help of human mask. Su et al.\cite{su2017pose} also use keypoints to guide division of human image so as to extract part-aligned local features. Different from these methods, the keypoints prediction in our method is separately conducted on different groups of feature channels, aiming to decouple the learned feature maps channel wise, which is a new way to implement local attention. We will show later that directly predict keypoints like most keypoint prediction methods does not work well in our ablation study~(Section~\ref{ablation_study_decouple}). Due to the decoupling learning of feature channels, our method improves robustness to occlusions and pose variance, thus better performance could be expected. As for the human mask prediction, our method uses it to guide holistic attention learning through back propagating, while previous works need to combine with a separate stream for predicting mask during inference. Finally, for existing ReID methods using these auxiliary information like human mask and keypoints, they use either existing detectors or separately trained ones, no co-training is involved as our do. They also have to change the original network both for training and test stages, and so the model size and computational cost in the inference stage are increased due to the added branches, such as~\cite{kalayeh2018human,song2018mask,wang2018mancs}.

Our work is also related to the multi-task learning in the perspective that we use two additional branches as implicit constraints to adjust the feature map learning in the backbone network. However, the most commonly used additional task for person ReID is adding classification loss~\cite{wang2018mancs,wang2018learning,zheng2019pyramidal}, which has the same purpose to ReID, i.e., distinguishing different persons. In this paper, we show two tasks with somehow contradicting purposes to ReID\footnote{Mask/keypoints prediction is to find common features among different people (unable to distinguish persons), while ReID targets at finding discriminative features for each person.} can be further used and lead to better performance.

\section{Our Approach}

\begin{figure*}[t]
	\vspace{-3mm}
	\begin{center}
		\begin{minipage}[t]{0.7\linewidth}
			\includegraphics[width=1.0\linewidth]{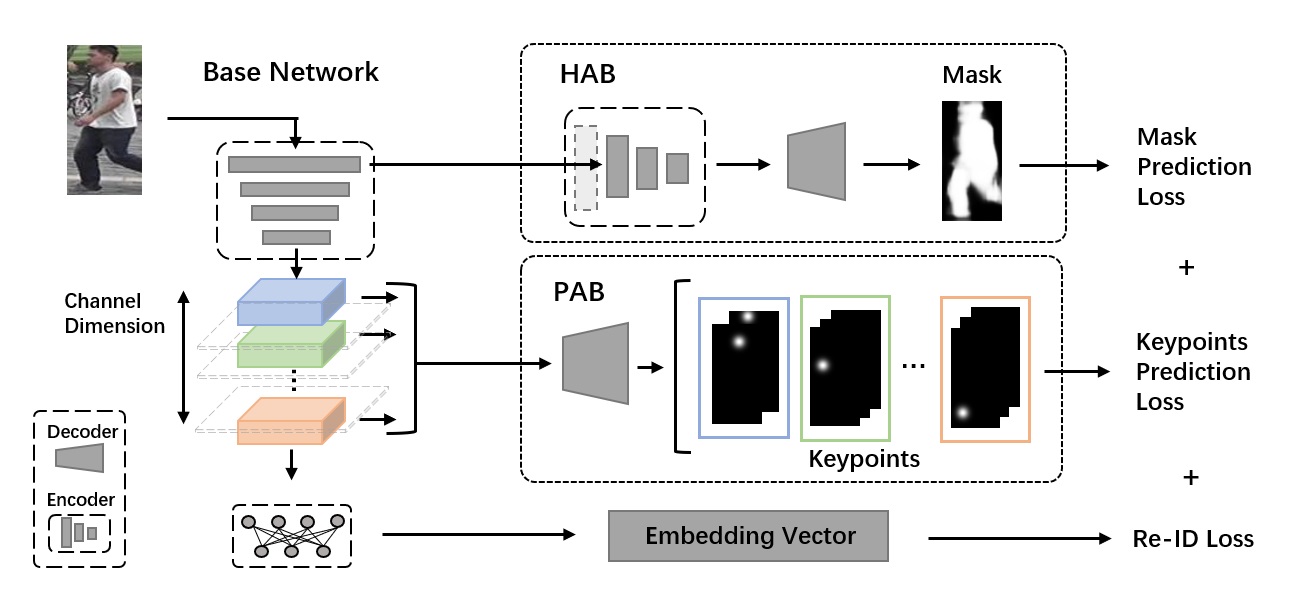}
		\end{minipage}
	\end{center}
	\vspace{-3mm}
	\caption{Structure of the proposed deep attention aware feature learning for person ReID. Our method adds two branches from the backbone network at training stage so as to guide the backbone being able to learn global and local attention aware features. The \emph{PAB} forces the separated groups of feature channels focus on predefined body parts by predicting their corresponding keypoints. The \emph{HAB} branch predicts the mask of a person and restricts the backbone network to focus on person bodies instead of background. PAB and HAB do not influence the testing stage.
	}
	\label{image:structure}
\end{figure*}

As shown in Fig.~\ref{image:structure}, our attention aware feature learning method can be applied to existing CNNs designed for person ReID~(whose structure is called as base network in this paper), by simply adding two branches from the backbone network during the training process. While for the inference using the trained network, the two added branches are removed and only the base network is used. Therefore, the proposed attention aware feature learning method could be considered as a general framework to adjust a ReID network by retraining with the two additional losses defined on the proposed branches respectively, while keeping the test network structure unchanged as the originial ReID network. On one hand, one branch named Holistic Attention Branch~(HAB) is designed for guiding the learned features being aware of the global human body in the clutter background, so as to make the backbone network pay more attention to the persons rather than the background. This is achieved by back propagating the supervision information about the human body mask through this branch to the backbone network. On the other hand, the Partical Attention Branch~(PAB) is proposed to make the learned features from backbone could be decoupled into different groups, each of which is capable of predicting several human body keypoints that are predefined according to their positions. In this way, the learned features are implicitly part-aligned, improving the robustness to occlusions and pose variations. As a result, our objective for learning feature embedding is
\begin{equation}\label{loss_all}
L = L_{r} + \lambda_{h} L_{h} + \lambda_{p} L_{p}
\end{equation}
where $L_{r}$ stands for a ReID loss~(e.g., the triplet loss for hard examples if we use TriNet~\cite{hermans2017defense} as the base network), $L_{h}$ is the loss computed on the HAB and $L_{p}$ is the one computed on PAB. $\lambda_{h}$ and $\lambda_{p}$ are two trade-off parameters. The details about the structures and the related losses are elaborated in the following. 

\subsection{Holistic Attention Feature Learning}
In order to make the learned features be able to focus more on the body part instead of the background, we use these features to predict the human body mask. The basic hypothesis for doing so is that the ability for generating the mask of a human body is highly correlated to the features' awareness of body part in the cluttered background. Therefore, if the mask can be predicted well, the features used for this task are considered being aware of body part in the background clutter. Inspired by this hypothesis, we introduce the Holistic Attention Branch to use the feature maps generated from the first convolution block~(i.e., the low-level features) as input, which are forwarded through some encoding and decoding layers to output the predicted human body mask. The structure of HAB is illustrated in the right top part of Fig.~\ref{image:structure}, where the encoder has the identical structure to the remaining part of the backbone network and the decoder consists of four deconvolution layers\cite{zeiler2010deconvolutional} and one $1\times 1$ convolution layer. The underlying reasons behind the design of encoding part of this branch are two folds. Firstly, the features extracted by CNNs gradually represent low-level to high-level ones. The low-level features are considered to be common ones for various tasks, while the high-level features are mostly task specific ones. For this reason, we need to build the mask prediction branch from the low-level features since human mask prediction is a different task to person ReID. Secondly, keeping the encoder has the identical structure to the backbone network can expect a good performance of human mask prediction as well as further imposes constraint on the shared low-level features to be general enough for different tasks. The detailed structure of the decoder in this branch is given in Table~\ref{parameter}. The loss $L_{h}$ related to this branch is
\begin{equation}\label{loss_mask}
L_{h} = \frac{1}{N} \sum\limits_{n=1}^{N} \|z_{n} - m_{n}\|_2
\end{equation}
where $N$ is the batch size, $z_{n}$ denotes the output of HAB for the $n$-th input, and $m_{n}$ is a binary image as the groundtruth body mask of the $n$-th input:
\begin{equation}\label{mask}
m_{n}(x',y') = \begin{cases}
1  & \text{if\ } (x',y') \text{\ is within a person body} \\
0  & \text{\ else}
\end{cases}
\end{equation}

\subsection{Partial Attention Feature Learning}
\label{PAB}

While the holistic attention feature learning could lead to a feature map focusing more on the human body out of the background clutter, local attention can be further helpful for person ReID. Lots of previous works have been proposed to explore the local features for ReID task. For example, PCB~\cite{sun2018beyond} dices image into horizon stripes and extracts local features from each parts. SPReID~\cite{kalayeh2018human}  uses a human semantic parsing branch to obtain local features.  PDC~\cite{su2017pose} uses keypoints explicitly to extract local features of different body part regions. All of these methods finally obtain feature representations by concatenating local features of different body parts~(and some methods even have to combine global features), thus inevitably resulting in a very high dimensional embedding vector for person ReID.  
On the contrary, we realize the partial attention feature learning by separating the feature maps extracted by the backbone network into different groups, each of which is trained to be responsible for predicting a specific group of keypoints. Compared to the previous works, our solution is more efficient as it does not increase the final feature embedding length. Moreover, since we use the same decoder to each group of feature maps for keypoint prediction, these groups are considered to be decoupled in the ideal case where each group can be used to predict the related keypoints perfectly. Although this can not be achieved in real case, the learning process is designed towards generating features with this property, thus our partial attention feature learning could also be considered as a kind of decoupled feature learning. The advantage of such decoupled feature learning is that it improves robustness to occlusions and pose variance. In case of occlusion happens, the disappeared body parts can only affect the corresponding feature channels, while other feature channels can still work well, thus the influence of partial occlusions to all feature channels could be limited to a small content.

\begin{figure}[tp]		 
	\vspace{-3mm}
	\begin{center}
		\begin{minipage}[t]{1.0\linewidth}
			\includegraphics[width=\linewidth]{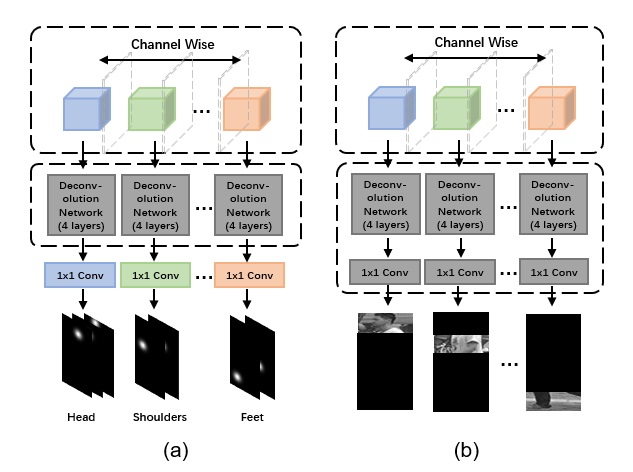}
		\end{minipage}
	\end{center}
	\vspace{-3mm}
	\caption{Structure of PAB branch. Keypoints or horizontal strips of different body parts could be predicted based on different feature channel groups. (a) is used for this work, and we will compare (a) and (b) in Section~\ref{ablation_study_decouple}.}
	\label{image:shareweights}
\end{figure}

Under this basic idea, the partial attention feature learning module takes the feature maps outputted by the backbone network as input, and manually separates them into several groups to predict different groups of keypoints. The learning procedure and the network structure are shown in the Fig.~\ref{image:structure} and Fig.~\ref{image:shareweights}(a). Supposing $a$ is the input image, $F$ is the mapping function of backbone network and $x=F(a)$ is the output feature maps of $a$. If we divide $x$ into $M$ groups, then $x=\{x_1,x_2,\cdots,x_M\}=\{F_1(a),F_2(a),\cdots,F_M(a)\}$, where $x_p=F_p(a)$ denotes the feature maps of the $p$-th group. Given $x_p$ as input, four deconvolution layers are followed to decoding the input as output feature maps with the size of input image, which is then convolved with a 1$\times$1 convolution to generate the keypoint prediction results~(i.e., in terms of heatmaps). Like other keypoint detection methods~\cite{chen2018cascaded}, $K$ keypoints are predicted with $K$ heatmaps respectively, each of which corresponds to one specific keypoint. Therefore, besides the sharing weights deconvolution layers, the 1$\times$1 convolution layers of different groups need to be independent since their role is to map the extracted features to different numbers of heatmaps.

Intuitively, different groups of feature maps can be used to predict different body parts, instead of the keypoints. For example, through image dicing, one can divide person image into horizontal stripes equally so that each stripe coarsely represents a part of human body. A straightforward of implementing the partial attention feature learning is to predict these divided horizontal stripes, given different groups of feature maps. However, such a method will be influenced by the background as the divided stripes contain background inevitably. What is worse, predicting the background as the body part will be harmful to the previously introduced holistic attention feature learning, whose role is to learning features focus on body part out of the background clutter. These are the reasons why we choose to predict the human body keypoints to realize the partial attention feature learning. In addition, since the PAB is built on feature maps after the HAB, it also benefits from the human body aware feature representations moduled by learning with HAB. This further emphasizes the importance of holistic attention learning.

Based on the above analysis, the proposed PAB is optimized according to the following regression loss for keypoint prediction:
\begin{equation}\label{loss_keypoints}
L_{p} = \frac{1}{N \cdot K} \sum\limits_{n=1}^{N}\sum\limits_{k=1}^{K}  \|p_{n,k} -  g_{n,k}\|_2
\end{equation}
where $N$ denotes the number of training samples in a batch, $K$ denotes the number of keypoints of a person; $k$ denotes the $k$-th output channel that corresponds to the $k$-th keypoint, $p$ and $g$ are the output and groundtruth respectively. For $K$ keypoints, there are $K$ output channels with the same size of input image, each of which predicts the location of a specific keypoint. Thus, $p_{n,k}$ denotes the $k$-th prediction channel for the $n$-th input image and $g_{n,k}$ is the groundtruth probability map of the $k$-th keypoint in the $n$-th image. Similar to the keypoint prediction methods~\cite{chen2018cascaded}, the groundtruth probability map of a keypoint is represented as a Gaussian distribution with the center in the position of keypoint.

\subsection{Implementations and Discussions}
\subsubsection{Network structure}
Although most of existing ReID networks can be used in our method as the base network, we implement our method on the basis of the widely used TriNet~\cite{hermans2017defense} without loss of generality.
Trinet\cite{hermans2017defense} consists of backbone network i.e. ResNet-50\cite{he2016deep} and 2 fully connected layers in the end, which is simple and concise. The parameters of decoders in HAB and PAB are listed in Table~\ref{parameter}. In addition, to demonstrate the universality of our method and how good performance can it achieve, we also implement our method on Bag of Tricks~\cite{luo2019bag} due to its state of the art performance.
The structure of Bag of Tricks is as same as Trinet except for a normalization layer and it is trained by optimize the triplet loss and cross entropy loss.
The two trade-off parameters in the learning object~(Eq.~(\ref{loss_all})) is set as $\lambda_{h}=\lambda_{p}=0.003$. The initial learning rate is 0.001. Note that the two additional branches are only used for network training. The original base network structure is used for testing. 

\begin{table}[htbp]
	\begin{center}
		\footnotesize
		\tabcolsep=3pt
		\newcommand{\tabincell}[2]{\begin{tabular}{@{}#1@{}}#2\end{tabular}}
		\begin{tabular}{ccccc}
			\hline
			layer & \#channels in & \#channels out & kernel size & stride \bigstrut\\
			\hline
			deconv 1 & \tabincell{c}{2048 (HAB)\bigstrut[t]\\ 341 (PAB)} & 64 & $3\times3$ & 2 \\
			deconv 2 & 64 & 64 & $3\times3$ & 2 \\
			deconv 3 & 64 & 64 & $3\times3$ & 2 \\
			deconv 4 & 64 & 64 & $3\times3$ & 2 \\
			$1\times1$ conv & 64 & \tabincell{c}{1 (HAB)\bigstrut[t]\\ keypoint groups (PAB)}  & $1\times1$ & 1 \bigstrut[b]\\
			\hline
		\end{tabular}
	\end{center}
	\caption{Parameters of deconvolution layers and $1\times1$ convolution layer in HAB and PAB.}
	\label{parameter}
\end{table}

\subsubsection{Weakly supervision of human body mask and keypoint} 
By adding two attention aware branches, our approach needs human mask and keypoint annotations to supervise the network training. However, manually labeling these annotations for the existing ReID dataset is impractical and labor expensive. Fortunately, there are many human mask detectors as well as keypoint detectors trained on the COCO dataset. We can freely use their detection results as weakly supervised information for these two branches. Although the human masks and keypoints generated in this way are not as accurate enough, our method can still benefit from the limited accurate annotations. In fact, our experiments show that the better the used detectors are, the higher ReID performance will our method achieve. Since the accuracy of these mask and keypoint detectors is still far from to reach the satisfactory level, it implies a big potential of using our method to improve the ReID performance. It is no doubt that our method will benefit from the development of these two techniques in the future. In this paper, we use \textit{CPN}\cite{chen2018cascaded} to generate keypoints of a person and ResNet-50 with 4-layer deconvolution head to generate human masks. Note that leveraging on existing detection results is not new in this paper, previous works~\cite{su2017pose} have been proposed to rely on keypoint alignment to improve ReID performance. Our method explores more on this direction and further improves previous results. The keypoint prediction in our method is only added in training stage and removed during inference. Thus, our method needs no extra computation cost on keypoints detection during inference.

\subsubsection{Keypoint grouping}
\begin{figure}[h]		 
	\vspace{-1mm}
	\begin{center}
		\begin{minipage}[t]{1.0\linewidth}
			\includegraphics[width=1\linewidth]{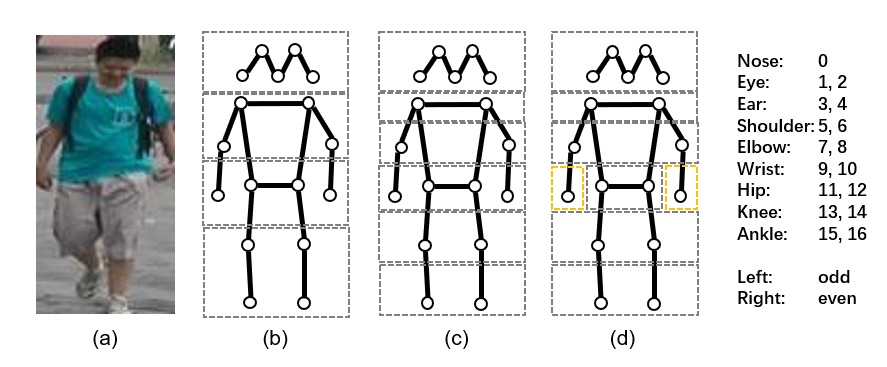}
		\end{minipage}
	\end{center}
	\vspace{-4mm}
	\caption{Illustration of different grouping schemes of human keypoints. (c) is used in this work, while other configurations are considered in our ablation study~( Section~\ref{ablation_study_decouple}).}
	\label{image:partation}
	\vspace{-5mm}
\end{figure}

In this paper, the keypoints of a person are divided into 6 groups according to their positions, as shown in Fig.~\ref{image:partation}. Accordingly, the feature maps used to predict the keypoints are also divided into 6 groups along the channel dimension, each of which is associated with a keypoint group. The output of \text{ResNet-50 Block-4} has 2048 channels and so each group has $\lfloor 2048/6 \rfloor = 341$ channels. The omitting 2 channels do not participate in keypoint prediction. We will show later in the ablation study that our method is insensitive to the specific way to group the keypoints~(cf. Table~\ref{ablation_keypoint_group}). What is important is to separate the feature channels into different groups for predicting keypoints respectively.

\section{Experiments}

\subsection{Datasets}
We use three widely used holistic datasets to validate the effectiveness of the proposed method, i.e. Market-1501 \cite{zheng2015scalable}, CUHK-03\cite{li2014deepreid}, DukeMTMC\cite{zheng2017unlabeled,ristani2016MTMC}. Market-1501 has 32668 annotated bounding boxes of 1501 identities collected by six cameras. There are 12936 images used for training. The query and gallery sets have 3368 and 19732 images respectively. CUHK-03 is a dataset consisting of 1467 identities for 13164 images. It is divided into two parts, CUHK-03(labeled) with manually labeled pedestrian bounding boxes and CUHK-03(detected) with DPM detected pedestrian bounding boxes. According to Zhong et al.\cite{zhong2017re}'s split, we divide the dataset into a training set and a testing set similar to Market-1501. DukeMTMC consists of 16522 training images of 702 identities, and 19889 images of the other 702 identities as testing set. The number of query images is 2228 and the number of gallery images is 17661.

Additionally, two partial datasets, i.e. Partial-REID \cite{zheng2015partial} and Partial-iLIDS \cite{zheng2011person} are used to evaluate the robustness of the proposed method to occlusions. Partial-REID has 60 identities, each of which has 5 partial images as query and 5 holistic images as gallery. Partial-iLIDS is a dataset consisting of 119 identities where one partial image is in query set and one holistic image is in gallery set for each person. Since there is no training data for Partial-REID and Partial-iLIDS, we test on these two partial datasets using the model trained on Market-1501.

\subsection{Results}

\begin{table*}[t]
	\begin{center}
		\newcommand{\tabincell}[2]{\begin{tabular}{@{}#1@{}}#2\end{tabular}}
		\resizebox{\textwidth}{!}{
			\begin{tabular}{c|ccccccccc}
				\hline
				\multirow{2}{*}{Model} &\multicolumn{2}{c}{Market-1501 (Single Query)} & \multicolumn{2}{c}{CUHK-03(labeled)}& \multicolumn{2}{c}{CUHK-03(detected)}&   
				\multicolumn{2}{c}{DukeMTMC-reID} \bigstrut\\
				\cline{2-9}    
				& mAP & Rank-1 & mAP & Rank-1 & mAP & Rank-1 & mAP & Rank-1 \bigstrut\\		
				\hline				
				
				HA-CNN~\cite{li2018harmonious} 
				& 75.70  & 91.20  & 41.00  & 44.40  & 38.60  & 41.70  & 63.80  & 80.50 \bigstrut[t]\\ 
				
				Mancs~\cite{wang2018mancs} 
				& 82.30  & 93.10  & 63.90  & 69.00  & 60.50  & 65.50 & 71.80  & 84.90 \\
				
				AACN~\cite{xu2018attention} 
				& 66.87  & 85.90 & -  & -  & -  & - & 59.25 & 76.84 \\	
				
				PDC~\cite{su2017pose} 
				& 63.41  & 84.14 & -  & -  & -  & - & - & - \\	
				
				PCB~\cite{sun2018beyond}
				& 81.06 & 93.80  & 57.50  & 63.70  & -  & -  & 69.20  & 83.30 \\
				
				MGCAM~\cite{song2018mask}
				& 74.33 & 83.79  & 50.21  & 50.14  & 46.87  & 46.71  & -  & - \\
				
				SPReID~\cite{kalayeh2018human} 
				& 83.36 & 93.68  & -  & -  & -  & -  & 73.34  & 85.95 \bigstrut[b]\\
				
				\hline
				
				TriNet~\cite{hermans2017defense}
				& 69.14 & 84.92  & 54.45  & 55.86  & 51.98  & 52.64  & 58.18  & 75.36 \bigstrut[t]\\
				
				DAAF-TriNet
				& 72.63 & 87.17  & 55.01  & 60.34  & 54.48  & 58.71  & 60.12 & 77.29 \\
				
				BoT~\cite{luo2019bag}
				& 85.90  & 94.50  & 60.90 & 63.30 & 58.00 & 59.10 & 76.40  & 86.40 \\
				
				DAAF-BoT 	& 87.90  & 95.10  & 67.60  & 69.00  & 63.10  & 64.90  & 77.90  & 87.90 \\
				
				BoT*
				& 94.20  & 95.40  & 77.10 & 74.40 & 73.40 & 70.40 & 89.10  & 90.30 \\
				
				DAAF-BoT*
				& 95.00  & 96.40  & 82.00  & 78.70  & 77.30  & 73.90  & 89.60  & 91.70 \bigstrut[b]\\
				
				\hline
			\end{tabular}	
		}
	\end{center}
	\caption{Comparison with state of the art on three large scale person ReID datasets. We apply the proposed deep attention aware feature learning~(DAAF) to two typical ReID models: TriNet~\cite{hermans2017defense} is one of the most influential works in deep learning based ReID, while Bag of Tricks~(BoT)~\cite{luo2019bag} with re-ranking~\cite{zhong2017re} achieves the state of the art performance on these benchmarks. * means re-ranking. Note that how DAAF improves existing methods. } 
	\label{table:result}
	\footnotesize 
\end{table*}

\subsubsection{Results on holistic datasets}
To demonstrate the effectiveness of the proposed deep attention aware feature learning~(DAAF), we apply it to two typical ReID models respectively. The first one is the TriNet~\cite{hermans2017defense}, which shows that the feature embedding for person ReID can be simply learned by a triplet loss using ResNet with hard examples mining. It is one of the most influential works in deep learning based ReID. The other one is the recently proposed Bag of Tricks~\cite{luo2019bag} which achieves the state of the art performance by collecting several practical training methods on ReID task and implanting a normalization layer to release the inconsistency between metric learning loss (i.e. triplet) and classification loss (i.e. cross entropy).
For comparison, we also report the results of some other methods from their original papers. These methods include three attention based methods~(HA-CNN~\cite{li2018harmonious}, Mancs~\cite{wang2018mancs}, AACN~\cite{xu2018attention}), two ReID methods based on local features~(PCB~\cite{li2018harmonious} and PDC~\cite{su2017pose}), and two methods  using human mask as additional information for ReID~(MGCAM~\cite{song2018mask} and SPReID~\cite{kalayeh2018human}). 

The quantitative results are reported in Table~\ref{table:result}. It is obvious that the proposed DAAF effectively improves over TriNet and Bag of Tricks in all cases. Especially, DAAF-Bag of Tricks with re-ranking achieves the state of the art performance in terms of both mAP and Rank-1 accuracy. When comparing to the base network, DAAF-TriNet improves over TriNet by 4.92\%-6.82\% for mAP and 3.36\%-7.16\% for Rank-1 accuracy on Market-1501, CUHK-03, and DukeMTMC-reID datasets. When using a better base network, DAAF-BoT achieves the state of the art performance on these benchmarks and finally DAAF-BoT* reaches the best results with re-ranking on all these tested datasets. 

The proposed DAAF is designed as additional objectives in the training stage in order to make the learned feature maps for ReID be aware of the person and the body parts. The greatest difference compared to other existing attention based methods lies in that this implicit manner of attention learning does not need to reweight a separate stream to the feature maps. Thus, our DAAF could be easily incorporated into other ReID networks. It could be seen from Table~\ref{table:result}, when added to Bag of Tracks, which already achieves the state of the art perfromance, our DAAF still could bring significant improvements on all these tested datasets, especially on the CUHK-03 dataset. Compared to existing methods based on local features or using human mask, our method takes both global and local attentions into consideration simultaneously. Particularly, the channel wise decoupling design in partial attention learning leads to the learned features for ReID more robust to pose variation and partial occlusion, which will be further verified in the following experiment results on partial datasets.

\begin{table}[h]
	\begin{center}
		\begin{tabular}{lcccc}
			\hline
			\multirow{2}[1]{*}{Model} & \multicolumn{2}{c}{Partial-iLIDS} & \multicolumn{2}{c}{Partial-REID} \bigstrut\\
			\cline{2-5}  & Rank-1 & Rank-3 & Rank-1 & Rank-3 \bigstrut\\
			\hline
			PCB~\cite{sun2018beyond} & 44.50 & 61.30 & 42.70 & 50.30\bigstrut[t]\\
			Trinet 		 			& 32.77 & 56.30 & 43.00 & 50.00\\
			\textbf{DAAF-Trinet} 	& \textbf{45.38} & \textbf{64.71} & \textbf{47.00} & \textbf{53.33}\\
			BoT 				& 51.30 & 63.90 & 57.30 & 65.30\\
			\textbf{DAAF-BoT} & \textbf{53.80} & \textbf{72.30} & \textbf{60.70} & \textbf{68.70} \bigstrut[b]\\
			\hline
		\end{tabular}
	\end{center}
	\caption{Results on two partial ReID datasets.}
	\label{partial_dataset_experiment}
\end{table}%

\subsubsection{Results on partial datasets} One advantage of our method is that the final feature embeddings are from those decoupled feature channels implicitly learned by PAB, therefore, if occlusion happens, only partial feature channels will be influenced, instead of all. As a result, our method is capable of dealing with partial occlusion in ReID. To show this point, we test our method on two additional partial ReID datasets, i.e., Partial-iLids and Partical-REID. For comparison, besides the base networks used in our method, we also report the results of PCB, which is a typical method based on local features. All the tested models are trained on Market-1501. Since the purpose of this paper is not for partial ReID, here we do not compare to those methods specially designed for partial ReID tasks that adopt different solutions to normal ReID tasks, which in turn has inferior ReID performance on general cases, such as the three benchmarks on Table~\ref{table:result}. The results shown in Table~\ref{partial_dataset_experiment} confirm that the proposed method is effective for handling the occlusion problem, with improved performance to the base network, either TriNet or Bag of Tricks. It also outperforms other local feature based methods such as PCB, even with a simple base network like TriNet.

\subsubsection{Visualization results of attention learning}
In order to visualize our attention learning, we use Grad-CAM (Gradient-weighted class activation maps)~\cite{selvaraju2017grad} to show the active part of our feature map. Slightly different from gradient-weighted class activation maps, we show the activation maps of feature embedding. Assume the embedding vector is $\textbf{x}=[x_1, ... ,x_N]^{T}$. According to the Grad-CAM\cite{selvaraju2017grad}, the gradient-weighted activation maps of $x_n$ can be presented by weighted sum of the feature maps:

\begin{equation}\label{feature_activation}
L_{Grad-CAM}^{n} = \frac{1}{K} \sum\limits_{k} Abs(F_{k}^{n} \odot A^{k}) 
\end{equation}
where $A^{k}$ means k-th channel of feature map, $K$ represents the number of channels, $\odot$ represent element-wise multiply and 

\begin{equation}\label{feature_weight}
F_{k}^{n} = \frac{\partial x_{n}}{\partial A^{k}}
\end{equation}
where the derivative of $x_n$ with respect to feature maps discribe the importance of each element from feature maps to the embedding component $x_n$. 

We add gradient-weighted activation maps of all channels to represent heatmap of holistic attention (Holistic Attention Map):

\begin{equation}\label{holist_attention_heatmap}
L_{HAM} = \frac{1}{N}\sum\limits_{n} L_{Grad-CAM}^{n}
\end{equation}

\begin{figure}[h]		 
	\begin{center}
		\begin{minipage}{0.9\linewidth}
			\includegraphics[width=1\linewidth]{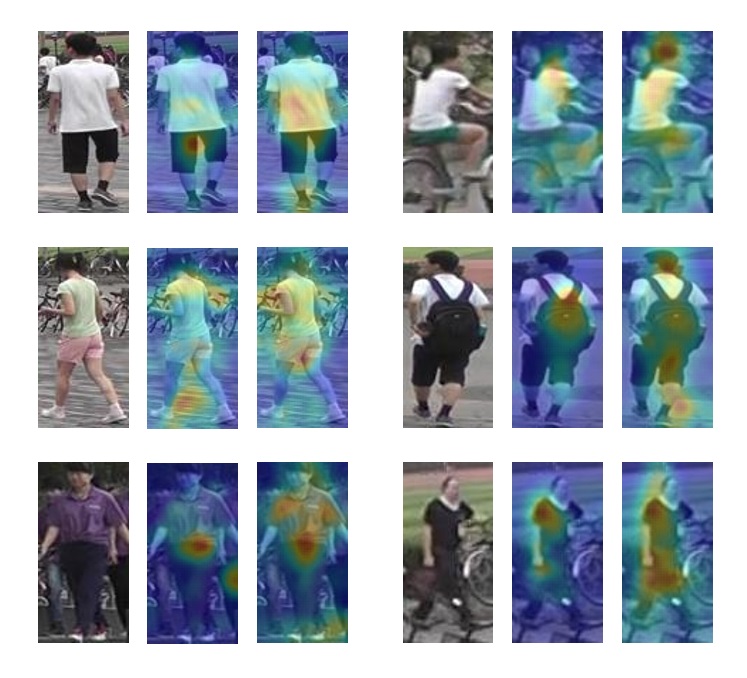}
		\end{minipage}
		
	\end{center}
	\caption{\textbf{Visualization of holistic attention feature learning.} For each group, the left is original image. The middle is the output activation maps of our baseline \textit{Trinet}\cite{hermans2017defense} without holistic attention learning, while the right is the output activation maps of \textit{Trinet}\cite{hermans2017defense} with holistic attention learning. }
	\label{fig:sp1}
\end{figure}

Using this method, we visualize the activation maps of our backbone network. In Equation~\ref{feature_activation}, we use final embedding vector as $\textbf{x}$ and take $\textbf{A}$ equal to \textit{ResNet conv1} feature map. As shown in Figure~\ref{fig:sp1}, the activation map of \textit{ResNet conv1} concentrate more on the whole person instead of background clutter or only a specific small part of body.

\begin{figure}[h]		 
	\begin{center}
		\begin{minipage}{0.9\linewidth}
			\includegraphics[width=1\linewidth]{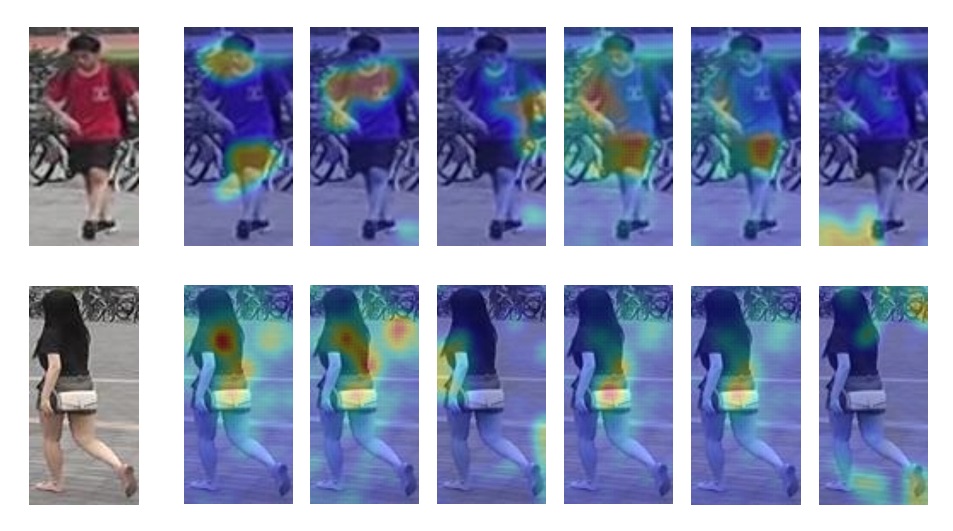}
		\end{minipage}
		
	\end{center}
	\caption{\textbf{Visualization of partial attention feature learning.} The left is original image. And from left to right, there are activation maps of 6 channel groups.}
	\label{fig:sp2}
\end{figure}

Similarly, instead of using the final embedding vector to visualize holistic attention, we use the vector generated by average pooling on different groups of the feature maps to visualize partial attention. The activation map of i-th group could be calculated by:
\begin{equation}\label{partial_attention_heatmap}
L_{PAM}^{i} = \frac{1}{Card(S_{i})}\sum\limits_{n\in S_{i}} L_{Grad-CAM}^{n}
\end{equation}
where $S_{i}$ is the set containing all serial number of channels in i-th group. As shown in Figure~\ref{fig:sp2}, through partial attention learning, different groups of channels could concentrate on different body parts.

\subsubsection{Learning results of HAB and PAB}
Additionally, although HAB and PAB are removed during inference, and the quality of their outputs is not related to ReID task directly. We still care about what HAB and PAB could output.

\begin{figure}[htbp]
	\begin{center}
		\begin{minipage}[t]{0.9\linewidth}
			\vspace{-1mm}
			\includegraphics[width=1.0\linewidth]{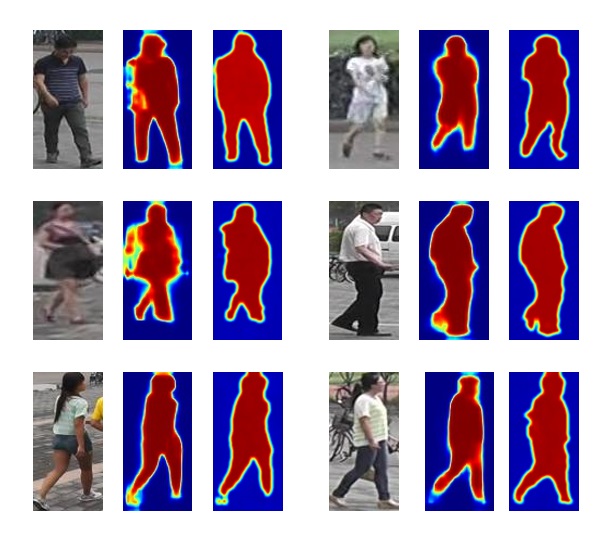}
		\end{minipage}
	\end{center}
	\vspace{-1mm}
	\caption{\textbf{Outputs of HAB during inference.} For each group, the left is image in testing set during inference, and the middle is the segmentation heatmap generated by the same mask detector as training sets'. The right is segmentation heatmap generated by HAB.
	}
	\label{image:mask_test}
\end{figure}

We add HAB to backbone network during inference and output the mask prediction results of HAB. To make comparison, we use the same mask detector which is used to obtain training labels to generate segmentation heatmap. As shown in Figure~\ref{image:mask_test}, the outputs of HAB are highly similar to the outputs of mask detector, which suggest that HAB obtain much useful information from the mask detector through using labels generated by the mask detector in training stage. The results also show that even though a small coefficient of mask loss is used to conduct weak supervision of human body, the learning results of HAB is not bad.

\begin{figure}[htbp]
	\begin{center}
		\begin{minipage}[t]{1.0\linewidth}
			\vspace{-1mm}
			\includegraphics[width=1.0\linewidth]{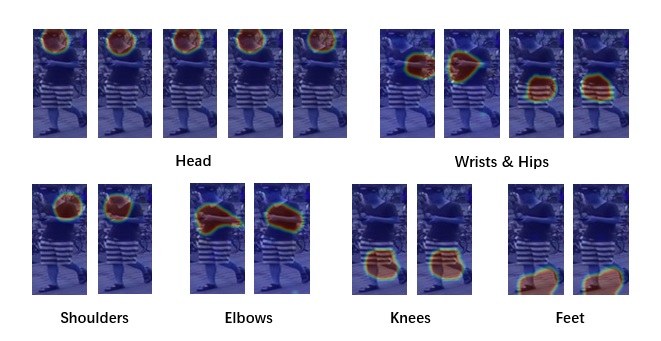}
		\end{minipage}
	\end{center}
	\vspace{-1mm}
	\caption{\textbf{Outputs of PAB during inference.} The heatmaps output by PAB which could be used for keypoints prediction. The high response region of heatmap is accurately corresponding to the position of keypoints.
	}
	\label{image:keypoint_test}
\end{figure}

Similarly, we add PAB to backbone network during inference and output the keypoints prediction results of PAB. Figure~\ref{image:keypoint_test} shows that the shape of keypoints prediction results varies from standard Gaussian Window, but the position of high response region is corresponding to the position of keypoints accurately.

\subsection{Ablation Study}

\subsubsection{The effectiveness of HAB and PAB}
We conduct experiment on Market-1501 to verify the effectiveness of the proposed HAB and PAB, whose results are shown in Table~\ref{ablation_modules}. To this end, we apply HAB and PAB separately to a base network~(TriNet) so as to observe how the performance changes. Since PAB aims to incorporate the partial attention about human keypoints into the feature learning process, the learned features will be aware of the global human body to some extent too. Therefore, compared to HAB that only enforces the feature learning towards perception of the human body, training with PAB is somewhat better as it enables the learned features not only being useful for predicting local information such as keypoints but also for the global human body. This analysis is consistent to the results shown in Table~\ref{ablation_modules}, where using PAB individually is better than using HAB. On the other hand, as other keypoint detection methods, PAB implements the keypoint prediction by heatmap regression where the groundtruth is generated by applying a Gaussian-window centered at the groundtruth keypoint. This kind of regression target inevitably contains some background areas, which can not be entirely suppressed by PAB alone. Therefore, HAB is required as a complementary to further alleviate the distraction of the background. Experimental results obtained when combining HAB and PAB together achieve the best results in Table~\ref{ablation_modules} validates this point.

\begin{table}[h]
	\begin{center}
		\begin{tabular}{cccc}
			\hline
			HAB & PAB & mAP & Rank-1 \bigstrut\\
			\hline
			\multicolumn{2}{c}{Base Network~(TriNet)} & 65.48 & 82.51 \bigstrut\\
			\hline
			\checkmark & & 68.26(+2.78) & 83.70(+1.19) \bigstrut[t]\\
			& \checkmark & 69.31(+3.83) & 84.41(+1.90) \\
			\checkmark & \checkmark & 70.40(+4.92) & 85.87(+3.36) \bigstrut[b]\\		
			\hline
		\end{tabular}
	\end{center}
	\caption{Ablation study of the PAB and HAB modules based on Market-1501 dataset. The numbers in the brackets are the improvements relative to the baseline.}
	\label{ablation_modules}
\end{table}

\subsubsection{Channel grouping for local attention}
\label{ablation_study_decouple}
In PAB, we propose a new method to learn the local attention aware features by channel decoupling. Each channel group is explicitly forced to learn features being responsible for a group of keypoints in a specific body part. In this way, the learning process will encourage different feature groups focus on different keypoints, while being less useful in predicting keypoints in other groups. Thus, achieving the decoupling of feature channels in term of keypoint prediction ability.  Alternatively, a typical technique used in the literature for learning features responsible for keypoint prediction is to take all the feature channels together as input to a decoder, and output $K$ probability maps, each of which corresponds to a specific keypoint. To show the advantage of our method, we implement this widely used technique to replace the proposed one in PAB. From the results reported in Table~\ref{ablation_decouple}, we can see that such a method to introduce local attention is not helpful. On the contrary, with our grouping strategy, the performance is significantly improved. This study validates that the boosted performance by the proposed PAB is not simply due to the incorporation of keypoint prediction in ReID, but more importantly, is because of the novel way~(decouple the feature channels) we proposed to use it. Note that in this experiment, we only apply PAB to the base network so as to reduce the influence of HAB for our analysis.

\begin{table}[h]
	\begin{center}
		\begin{tabular}{ccccc}
			\hline
			Methods & with Group & w/o Group  & mAP & Rank-1 \bigstrut\\
			\hline
			keypoint & & \checkmark & +0.92 & +0.03 \bigstrut[t]\\
			part image & \checkmark & & +2.74 & +1.66 \\
			keypoint  & \checkmark & & +3.83 & +1.9 \bigstrut[b]\\
			\hline
		\end{tabular}
	\end{center}
	\caption{The effectiveness of channel wise decoupling in PAB. The results are reported on the Market-1501 dataset. The numbers are the relative increase/decrease compared to the base network, i.e., TriNet. No HAB is used in order to focus on different choices of PAB. 
	}
	\label{ablation_decouple}
\end{table}

To further show the importance of decoupling feature channels, we conduct another experiment on PAB by replacing the keypoint prediction with the part image prediction. Specifically, we divide the input image into 6 equal parts horizontally as PCB~\cite{sun2018beyond} does. As shown in Fig.~\ref{image:shareweights}(b), the groundtruth consists of blank area and specific image part, in which only 1/6 image could be seen. Then we separate the output feature channels of backbone into 6 groups, each of which is responsible for predicting a corresponding body part image. As shown in Table~\ref{ablation_decouple}, even with such a simple supervision, PAB still works well and improves the baseline substantially. Compared to using keypoint prediction as supervision information, predicting the partial body image is inferior because the horizontal body part image inevitably contains a larger portion of background than keypoint and the part appearance is more complex than the heatmap of keypoints. Therefore, although the feature channel decoupling is key to the performance improvement in PAB, it can still benefit from a well designed task relying on local features, such as the keypoint prediction used in this paper. 
How to incorporate a better task for realizing the local attention is beyond the scope of this paper, and will be studied as our future work.

Finally, although PAB relies on keypoint grouping to implement the channel wise decoupling, the specific way of keypoint grouping only has a limited influence on its performance according to our experiments~(see Table~\ref{ablation_keypoint_group} for the results of different grouping schemes shown in Fig.~\ref{image:partation}). The possible reason could be that different keypoints of a human body are mostly related to each other, it is hard to manually find a good grouping of them so as to make different groups are potentially unrelated, in which case the performance may benefit most from the proposed decoupling learning. In common cases, there is little difference when there are correlations among different groups. Among the tested grouping schemes, as shown in Table~\ref{ablation_keypoint_group}, the best performance is achieved when using 6 parts which is closely followed by utilizing 7 parts. This results is relatively consistent to some other methods which use similar body part grouping schemes \cite{zhao2017deeply,sun2018beyond}.

\begin{table}[htbp]
	\begin{center}
		\newcommand{\tabincell}[2]{\begin{tabular}{@{}#1@{}}#2\end{tabular}}
		\begin{tabular}{cccc}
			\hline
			Num  & mAP & Rank-1  & Grouping Scheme \bigstrut\\
			\hline
			1 & 66.40 & 82.54 & all\\
			4 & 69.31 & 84.41 & Fig.~\ref{image:partation}(b)\\
			\bf{6} & \bf{70.40} & \bf{85.87} & Fig.~\ref{image:partation}(c) \\
			7 & 70.04 & 85.72 & Fig.~\ref{image:partation}(d)\\
			17 & 69.82 & 85.12 & each\\		
			\hline
		\end{tabular}
	\end{center}
	\caption{Performance on Market-1501 for different keypoint grouping schemes shown in Fig.~\ref{image:partation}. "all" refers to use all keypoints together as one group, while "each" is to use each keypoint separately for one group.}
	\label{ablation_keypoint_group}
\end{table}

\subsubsection{Weights sharing mechanism}
Shared-weights deconvolution layers are incorporated into PAB to conduct channel grouping, which means six groups of keypoints prediction use the same decoder. The reason why we don't use 6 independent decoders is that decoders should not be related to the group of keypoints or they may have partial information of human body. The goal of partial attention learning is to let different channels of feature map focus on different parts rather than let different decoders focus on different parts. Therefore, using the same decoder for all channel groups will guarantee the decoder not related to a specific body part, which could let different channels of feature maps focus more on corresponding body parts. For comparison, we use 6 independent branches to conduct the keypoints or diced image prediction. The results are shown in Table~\ref{ablation_weights_sharing}.

\begin{table}[h]
	\begin{center}
		\begin{tabular}{ccccc}
			\hline
			Methods & same & independent & mAP & Rank-1 \bigstrut\\
			\hline
			keypoint 	& \checkmark & & +3.83 & +1.90 \bigstrut[t]\\
			keypoint 	& & \checkmark & +1.35 & -0.03 \\
			part image 	& \checkmark & & +2.74 & +1.66 \\
			part image  & & \checkmark & +1.33 & +0.77 \bigstrut[b]\\
			\hline
		\end{tabular}
	\end{center}
	\caption{We use 6 independent decoders to predict keypoints or part images in comparsion with the method using the same decoder. No HAB is used in order to explore PAB design.}
	\label{ablation_weights_sharing}
\end{table}

As shown in Table~\ref{ablation_weights_sharing}, partial attention learning using 6 independent branches performs worse than using the same one. The results based on keypoints and based on part images are consistent, which suggest that using the same decoder is essential to decoupled feature learning. 

\begin{figure*}[htbp]
	\begin{center}
		\begin{minipage}[t]{\linewidth}
			\vspace{-1mm}
			\includegraphics[width=1.0\linewidth]{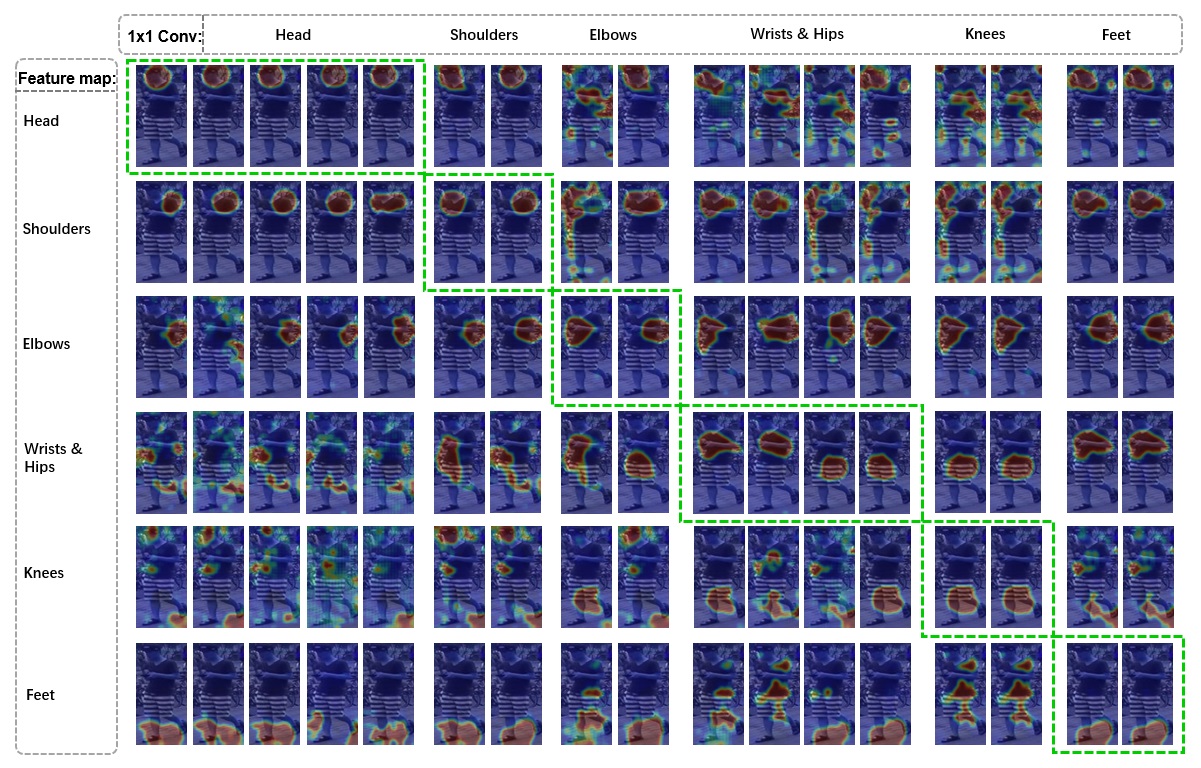}
		\end{minipage}
	\end{center}
	\vspace{-1mm}
	\caption{\textbf{Results of 1x1 conv shuffle.} In each row, a fixed set of feature maps is combined with different 1x1 convolution layers. The diagonal of the image represents feature map and 1x1 convolution layer are matched. The outputs are mainly determined by inputs feature maps instead of 1x1 convolution layers.
	}
	\label{image:shuffle_outputs}
\end{figure*}

\subsubsection{1x1 conv shuffle in PAB}

\begin{figure}[h]
	\begin{center}
		\begin{minipage}[t]{\linewidth}
			\vspace{-1mm}
			\includegraphics[width=1.0\linewidth]{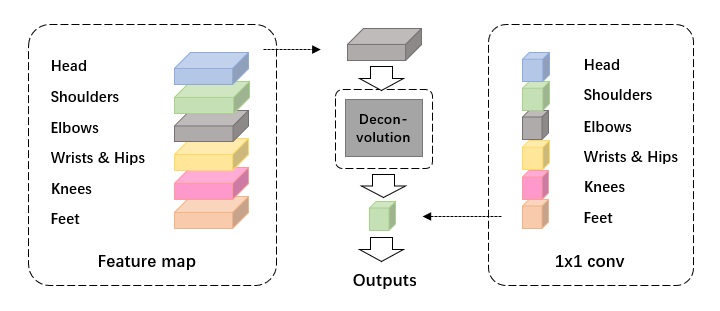}
		\end{minipage}
	\end{center}
	\vspace{-1mm}
	\caption{\textbf{1x1 conv shuffle in PAB.} We combine the 1x1 convolution layer of different groups with different feature maps to conduct keypoints prediction. All 1x1 convolution layer and feature maps are trained normally, and we only shuffle them during inference.
	}
	\label{image:shuffle}
\end{figure}

Because of different number of keypoints in each group, the 1x1 convolution layers are different while the deconvolution layers for each group are same. We will show that, to a great extent, 1x1 convolution layers contains little part-related information. We conduct 1x1 convolution layers shuffle experiments. As shown in Figure~\ref{image:shuffle}, we combine different channel groups of feature map with different 1x1 convolution layers during inference. 
If each group of the output heatmaps corresponds to the 1x1 convolution layers no matter which group of feature maps input, 1x1 convolution layers instead of the feature maps are responsible for predicting the keypoints of different body parts. This suggests each group of feature maps doesn't well focused on the corresponding body part. On the contrary, if each group of the output heatmaps corresponds to the input feature maps no matter which group of 1x1 convolution layers is used, the 1x1 convolution would include little information of body parts.

The results are shown in Figure~\ref{image:shuffle_outputs}. The results show that the outputs are mainly determined by inputs feature maps instead of 1x1 convolution layers, which suggests that the feature are possibly highly decoupled on channel dimension.

\section{Conclusion}
In this paper, we propose an attention aware feature learning method for person re-identification. The proposed method consists of a partial attention branch~(PAB) and a holistic attention branch (HAB) that are jointly optimized with the base re-identification feature extractor. Since the two branches are built on the backbone network, it does not introduce additional structure for ReID feature extraction. Therefore, our method is able to maintain the same inference time as the original network, while most previous works realize the attention mechanism at the cost of higher inference time. Experimental results show that the proposed feature learning method is general enough by integrating it into two different base networks and consistent performance improvement is observed. With a strong base network such as Bag of Tricks proposed recently, our method achieves the state of the art performance on various ReID benchmarks. In addition, due to the channel wise decoupling features learned through PAB, the proposed method also works well on ReID tasks with partial occlusion.


{\small
\bibliographystyle{ieee_fullname}
\bibliography{mybibfile}
}

\end{document}